\documentclass[twocolumn]{el-author}

\usepackage{textcomp}
\usepackage{float}
\usepackage[misc]{ifsym}
\usepackage{amsmath}
\usepackage{algpseudocode}

\begin{document}

\title{{SAR} Image Despeckling Using Quadratic-Linear Approximated  $\ell_1$-Norm}

\author{Fatih Nar \textsuperscript{\Letter}}

\abstract{Speckle noise, inherent in synthetic aperture radar (SAR) images, degrades the performance of the various SAR image analysis tasks. 
Thus, speckle noise reduction is a critical preprocessing step for smoothing homogeneous regions while preserving details. 
This letter proposes a variational despeckling approach where $\ell_1$-norm total variation regularization term is approximated in a quadratic and linear manner to increase accuracy while decreasing the computation time.
Despeckling performance and computational efficiency of the proposed method are shown using synthetic and real-world SAR images.}

\maketitle

\section{Introduction}
Synthetic aperture radar (SAR) is a microwave sensor system that allows acquiring high-resolution images at day or night, and almost in all weather conditions. 
However, speckle noise degrades SAR image quality and causes difficulties for various image analysis tasks (i.e. edge detection, change detection, segmentation) \cite{argenti2013} \cite{ozcan2015sdd}. 

In this letter, variational approach for despeckling is employed due to its excellent performance in various image processing tasks \cite{ozcan2015sdd}.
In the literature, an anisotropic diffusion process for edge preserving noise reduction is proposed by Perona and Malik \cite{perona1990}.
Then, variational noise reduction, ROF model, is proposed in \cite{rudin1992rof} where the diffusion process is controlled with a data fidelity term. 
Afterwards, various despeckling methods are proposed for SAR images such as speckle reducing anisotropic diffusion (SRAD) \cite{yu2002srad}, improved anisotropic diffusion \cite{fabbrini2013}, and sparsity-driven despeckling (SDD) \cite{ozcan2015sdd}. 
In this study, the approximation of the TV regularization term in SDD is improved to increase despeckling accuracy while reducing execution time. 

\section{Proposed method}
Speckle reduction for the SAR image is defined as the minimization of the following variational cost function:
\begin{equation}
	\label{TV_denoising_cost_function}
	J(f) =  \frac{1}{2N} \sum_{p=1}^{N} (f_p - g_p)^2 + \lambda|(\nabla{f})_p|  
\end{equation}
where $g$ is observed speckled image, $f$ is the desired despeckled image, $N$ is the pixel count, $p$ is the pixel index number, $\lambda$ is a positive value determining smoothing level, and $\nabla$ is the gradient operator. 
In the cost function, the data fidelity term ensures $f$ stays similar to $g$ in $\ell_2$-norm manner and total variation (TV) regularization term implies penalty on the changes in image gradients in $\ell_1$-norm manner.

\begin{figure}[h]
\centering{\includegraphics[width=75mm]{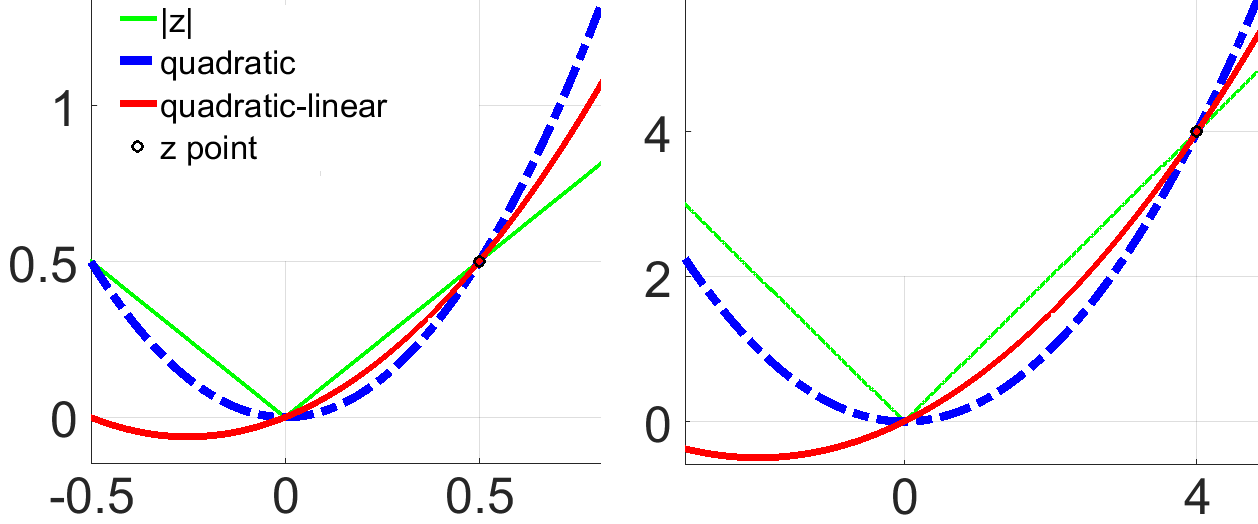}}
\vspace*{-3 mm}
\caption{Quadratic and QL ($\alpha=0.5$) $\ell_1$-norm approximations at $0.5$ and $4$}
\label{QL_L1_appoximation_figure}
\end{figure}

Although $\ell_1$-norm TV regularization preserves details, its efficient minimization is difficult since it is not differentiable. SDD \cite{ozcan2015sdd} proposed to approximate the non-differentiable term quadratically as below:
\begin{equation}
	\label{quadratic_approximation}
	|z| \approx (|\hat{z}| + \varepsilon) ^{-1} {z^2}
\end{equation}
where $\hat{z}$ is a proxy constant for $z$ and $\varepsilon$ is a small positive constant. 
Accuracy of the approximation increases as $\varepsilon$ gets closer to $0$. 
In this study, this quadratic approximation is further improved by combining it with a linear approximation as given in equation (\ref{quadratic_linear_approximation}).
\begin{equation}
	\label{quadratic_linear_approximation}
	|z| \approx (1 - \alpha)(|\hat{z}| + \varepsilon) ^{-1} {z^2} + \alpha sgn(\hat{z})z
\end{equation}
where $0 \leqslant \alpha \leqslant 1$, $sgn(.)$ is the signum function, and $sgn(\hat{z})z$ is the linear approximation of $|z|$. 
Equation (\ref{quadratic_linear_approximation}) is convex combination of quadratic and linear approximations and is accurate around $\hat{z}$ (see Fig. \ref{QL_L1_appoximation_figure}). 
As $z$ goes to $0$, linear term vanishes and quadratic-linear (QL) approximation becomes quadratic around $0$ which also avoids staircase artifacts.

If we define $|(\nabla{f})_p|$  as $|(\partial_x{f})_p| + |(\partial_y{f})_p|$ for a 2D SAR image and use the QL approximation given in equation (\ref{quadratic_linear_approximation}) then the cost function in equation (\ref{TV_denoising_cost_function}) becomes as below:
\begin{equation}
\begin{aligned}
	\label{TV_denoising_cost_function_approximated}
	J^{(n)}(f) =  \frac{1}{2N} & \sum_{p=1}^{N} 
			 (f_p - g_p)^2 + (f_p - \hat{f}_p)^2  \\ 
			 & + \lambda [ (1 - \alpha)(w_{x,p} {(\partial_x{f})_p^2}  + w_{y,p} {(\partial_y{f})_p^2})  \\
			 & ~~~~~~~ + \alpha (s_{x,p} {(\partial_x{f})_p} + s_{y,p} {(\partial_y{f})_p}) ]
\end{aligned}			
\end{equation}
where $n$ is the iteration number,
$\hat{f}_p$ is a proxy constant for $f_p$,  
$(f_p - \hat{f}_p)^2$ is a new regularization term for forcing $f_p$ stays close to $\hat{f}_p$ since QL approximation is only accurate around $\hat{f}_p$, 
$w_{x,p}=(|(\partial_x{\hat{f}})_p| + \varepsilon) ^{-1}$, 
$s_{x,p}=sgn((\partial_x{\hat{f}})_p)$, 
and $w_{y,p}$ and $s_{y,p}$ are defined correspondingly.
Superscript $n$ in $J^{(n)}(f)$ shows that cost function must be minimized in an iterative manner due to employed QL approximation.

Equation (\ref{TV_denoising_cost_function_approximated}) can be represented in matrix-vector form as below:
\begin{equation}
\begin{aligned}
	\label{TV_denoising_cost_function_matrix_vector}
	J^{(n)}(f) =  \frac{1}{2N} & \Big(
			 (v_f - v_g)^{\top}(v_f - v_g) + (v_f - v_{\hat{f}})^{\top}(v_f - v_{\hat{f}})  \\ 
			 & + \lambda [ (1 - \alpha)(v_f^{\top} C_x^{\top} W_x C_x v_f + v_f^{\top} C_y^{\top} W_y C_y v_f)  \\
			 & ~~~~~~~ + \alpha (s_x^{\top} C_x v_f + s_y^{\top} C_y v_f) ]   \Big)
\end{aligned}			
\end{equation}
where $v_g$, $v_f$, $v_{\hat{f}}$, $s_x$, $s_y$ are vector forms of $g_p$, $f_p$, $\hat{f}_p$, $s_{x,p}$, $s_{y,p}$,
and $W_x$, $W_y$ are diagonal matrix form of $w_{x,p}$, $w_{y,p}$,
and $C_x$, $C_y$ are the Toeplitz matrices as the forward difference gradient operators where derivatives are zero at the right and bottom boundaries respectively.

Equation (\ref{TV_denoising_cost_function_matrix_vector})  is strictly convex and differentiable; thus, one can take its derivative with respect to $v_f$ and equalize it to zero to obtain its minimum which leads to a linear system as given below:
\begin{equation}
\begin{aligned}
	\label{TV_denoising_linear_system}
	A v_f^{(n+1)} = b
\end{aligned}			
\end{equation}
where $A = 2I+ \lambda (1 - \alpha) (C_x^{\top} W_x C_x + C_y^{\top} W_y C_y)$, $I$ is identity matrix, and $b = v_g + v_{\hat{f}} - \lambda (\alpha / 2) (C_x^{\top} s_x + C_y^{\top} s_y) $.  
Iteration number is $n$ for the $A$, $W_x$, $W_y$, $b$, $v_f$,  $v_{\hat{f}}$, $s_x$, and $s_y$ unless it is explicitly stated.

Pseudo-code of the proposed method is given in algorithm \ref{QL_L1_PseudoCode} where implementation of all the steps are easy and computationally cheap, except for solving the linear system in line 11.
To obtain computational efficiency in line 11, preconditioned conjugate gradient (PCG) with incomplete Cholesky preconditioner (ICP) is used where maximum PCG iteration is set to $10^2$ and convergence tolerance is set to $10^{-2}$.
Note that, all the matrices ($C_x$, $C_y$, $W_x$, $W_y$, $A$) in algorithm \ref{QL_L1_PseudoCode} are sparse.
\begin{algorithm}
{\fontsize{7pt}{7pt}\selectfont
\caption{Quadratic-Linear Approximated $\ell_1$-norm Despeckling}
\label{QL_L1_PseudoCode}
\begin{algorithmic}[1]
	\Procedure{SDD-QL}{$g, \lambda, \varepsilon, \alpha, n_{max}$}					\Comment{SDD with QL (SDD-QL)}
	\State $v_f \gets v_g \gets g$												\Comment{$g$ is observed speckled image}
	\For{n=1} {$n_{max}$}														\Comment{$n_{max}$ is the maximum iteration}
		\State $v_{\hat{f}} \gets v_f$
		
		\State $W_x \gets [\textit{diag}(|C_x v_{\hat{f}}| + \varepsilon)]^{-1}$		\Comment{$C_x v_f$ is $x$-derivative of $v_f$}
		\State $W_y \gets [\textit{diag}(|C_y v_{\hat{f}}| + \varepsilon)]^{-1}$		\Comment{$C_y v_f$ is $y$-derivative of $v_f$}

		\State $s_x \gets \textit{sgn}(C_x v_{\hat{f}})$							\Comment{signum of $x$-derivative of $v_f$}
		\State $s_y \gets \textit{sgn}(C_y v_{\hat{f}})$							\Comment{signum of $y$-derivative of $v_f$}

		\State $A \gets 2I+ \lambda (1 - \alpha) (C_x^{\top} W_x C_x + C_y^{\top} W_y C_y)$
		\State $b \gets v_g + v_{\hat{f}} - \lambda (\alpha / 2) (C_x^{\top} s_x + C_y^{\top} s_y)$
		
		\State solve $A v_f = b$ 													\Comment{solve equation (\ref{TV_denoising_linear_system}) to find $v_f$ for the next iteration}
	\EndFor
	\State $f \gets v_f$
\State \textbf{return} $f$															\Comment{return despeckled image}
\EndProcedure
\end{algorithmic}
}
\end{algorithm}

As $\alpha$ gets closer to $1$, $A$ become more diagonally dominant and efficiency for solving the linear system increases.
However, in that case diffusion process is calculated in a local manner which leads to tiny dithering artifacts in the result. 
For $\alpha = 1$, $A$ becomes diagonal and solution of the linear system in equation (\ref{TV_denoising_linear_system}) becomes very efficient but more outer iterations ($n_{max}$) are required. 
For $\alpha < 1$, $A$ becomes a positive definite and sparse 5-point Laplacian matrix which can be solved with an efficient iterative solver such as PCG. 
As $\alpha$ gets closer to $0$, $A$ become less diagonally dominant; therefore, efficiency for solving the linear system decreases while diffusion becomes more global and only few outer iterations ($n_{max} =5$) are required. 
In SDD-QL, best accuracy and computational efficiency is achieved when $\alpha$ is around $0.5$. 

\section{Results and analysis}
In this section, SDD with QL (SDD-QL), is analyzed qualitatively and quantitatively to show its despeckling accuracy and computational efficiency. 
SDD and SDD-QL are both developed in C++ using the coding optimizations given in \cite{ozcan2015sdd} and compiled as 64 bit executables.
In all the experiments, 
	(a) Intel i7-6700K 4 GHz CPU is used as hardware, 
	(b) TerraSAR-X sample SAR image of India Visakhapatnam port (spot-mode, 16 bit, VV polarization, resolution $\approx$ 1 meter, number of looks $\approx$ 1) is used as test image, and
	(c) $\lambda=100$, $\varepsilon=10^{-2}$, $\alpha=0.5$, and $n_{max}=5$ are default parameters.

\begin{figure}[h]
\centering {
	\includegraphics[width=0.45\textwidth]{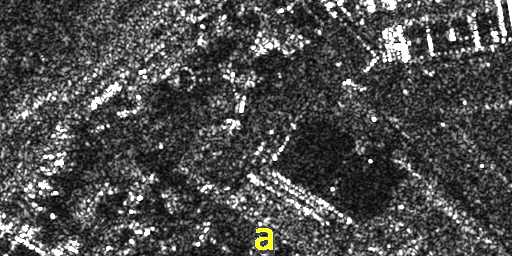} \\
	\includegraphics[width=0.45\textwidth]{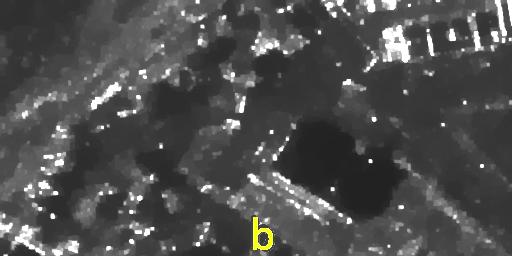} \\
	\includegraphics[width=0.45\textwidth]{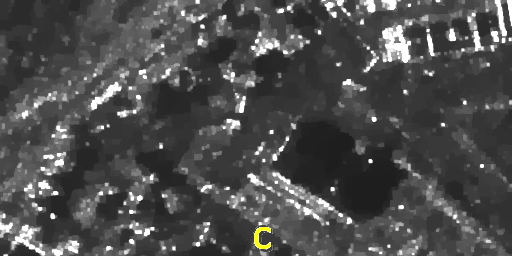} \\
}
\vspace*{-3 mm}
\caption{SDD vs SDD-QL: a) SAR image, b) SDD result , c) SDD-QL result}
\label{SDDversusSDDQL}
\end{figure}

As seen in Figure \ref{SDDversusSDDQL}, SDD and SDD-QL produce similar despeckling results since SDD-QL is a variant of SDD. 
However, SDD-QL preserves reflectivity levels in each region better due to the applied improvements on SDD while homogeneous regions are smoothed equivalently. 
Better reflectivity preservation of the regions leads to better preservation of details such as point scatterers and edges.
Improvements obtained by SDD-QL can be observed with a closer investigation in Figure \ref{SDDversusSDDQL}.

\begin{figure}[h]
\centering {
	\includegraphics[width=0.155\textwidth]{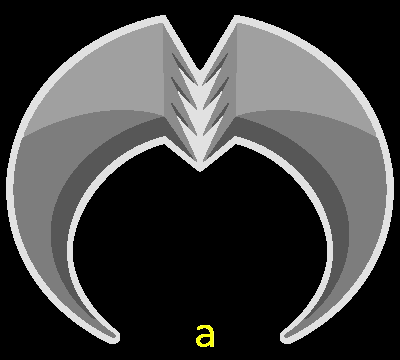}
	\includegraphics[width=0.155\textwidth]{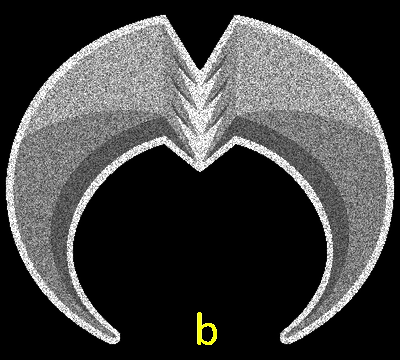}
	\includegraphics[width=0.155\textwidth]{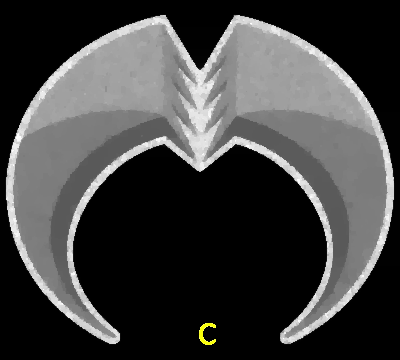}
}
\vspace{0pt}
\vspace*{-3 mm}
\caption{Synthetic data: a) KFAU logo, b) 1-look SAR image, c) SDD-QL result}
\label{SyntheticImageExample}
\end{figure}

\begin{figure}[h]
\centering{\includegraphics[width=85mm]{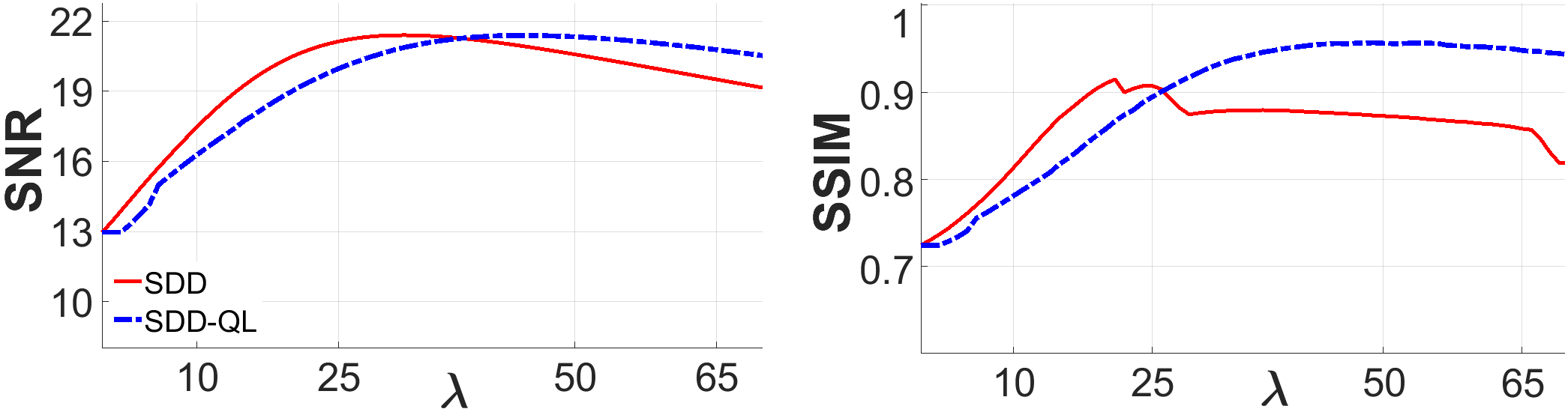}}
\vspace{0pt}
\vspace*{-3 mm}
\caption{SDD vs SDD-QL SNR and SSIM comparison ($\varepsilon = 10^{-4}$)}
\label{SNR_SSIM_Comparison}
\end{figure}

In Figure \ref{SyntheticImageExample}, a synthetically generated SAR image of KFAU logo and its SDD-QL despeckling result are shown. 
For this synthetic data, Figure \ref{SNR_SSIM_Comparison} shows signal to noise ratio (SNR) and structural similarity (SSIM) index values for different $\lambda$ parameters.
In this experiment, 
	speckled image has SNR=$12.969$dB with SSIM=$0.725$, 
	best result of SDD-QL has SNR=$21.395$dB with SSIM=$0.956$, 
	and best result of the SDD has SNR=$21.133$dB with SSIM=$0.907$. 
As seen in Figure \ref{SNR_SSIM_Comparison}, both methods achieve similar level of SNR while SDD-QL achieves higher value of SSIM which shows that SDD-QL preserves edges better than SDD.
~

\begin{figure}[h]
\centering{\includegraphics[width=80mm]{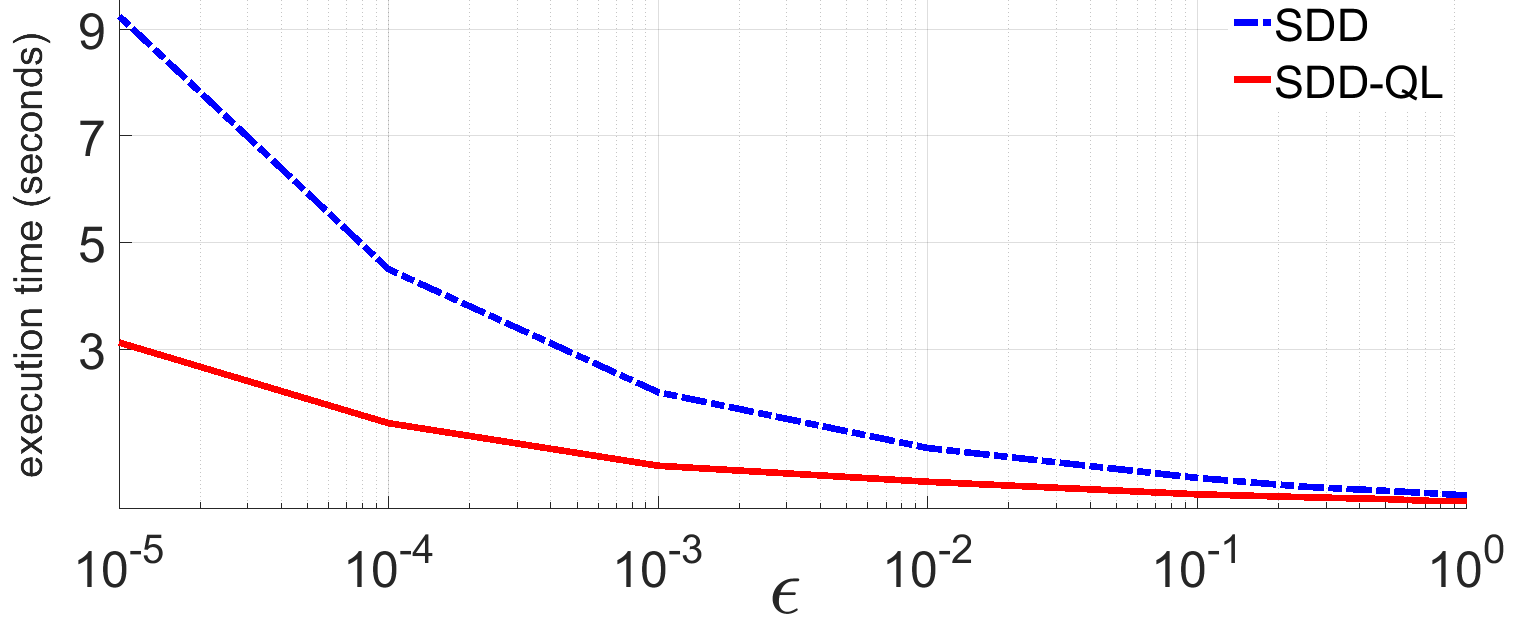}}
\vspace{0pt}
\vspace*{-3 mm}
\caption{SDD vs SDD-QL execution time (512x512 SAR image)}
\label{executionTimeComparison}
\end{figure}

Quadratic part of the QL approximation hence QL approximation itself gets better as $\varepsilon$ gets smaller. 
However, $A$ matrix becomes more ill-conditioned as the $\varepsilon$ gets smaller; thus, solving the linear system in equation (\ref{TV_denoising_linear_system}) gets longer.
Note that, SDD and SDD-QL use PCG with ICP to solve the linear system which leads to significantly faster computation compared to no preconditioning.
Even so, like the other preconditioners, ICP also sacrifices the preconditioning performance to obtain efficient construction of the preconditioner to decrease the overall computation in PCG.
SDD-QL method produces better conditioned $A$ matrix compared to the one produced by SDD; thus, SDD-QL is faster. 
For general despeckling tasks, $\varepsilon$ can be set as $10^{-1}$ where SDD-QL is 2 times faster compared to SDD; and for very accurate despeckling tasks, $\varepsilon$ can be set as $10^{-5}$ where SDD-QL is almost 3 times faster compared to SDD (see Fig. \ref{executionTimeComparison}).
For $\varepsilon = 10^{-1}$, SDD-QL despeckles a 512x512 pixels SAR image in 0.28 second in single thread and despeckles a 13312x8192 pixels SAR image in 23.20 seconds with 8 threads.

\section{Conclusion}
In this letter, approximation of the TV regularization term in SDD method is improved by fusion of a quadratic and linear approximators.
Presented quadratic-linear approximator is derived for $\ell_1$-norm, but it can be easily extended to other norms that provides sparsity.
Experiments show that, proposed method leads to more accurate despeckling with up to 3 times faster execution times comparing to SDD even though SDD already uses satisfactory $\ell_1$-norm approximation and an efficient numerical schema.

\vskip3pt
\ack{Author would like to thank Atilla Ozgur, Osman Erman Okman, and Mujdat Cetin for their useful suggestions.}

\vskip3pt

\noindent Fatih Nar (\textit{Konya Food and Agriculture University (KFAU), Turkey})
\vskip3pt

\noindent \scriptsize{\Letter} ~ \small E-mail: fatih.nar@gidatarim.edu.tr

\end{document}